\documentclass[runningheads]{llncs}
\usepackage[T1]{fontenc}
\usepackage{graphicx}

\usepackage[utf8]{inputenc} %
\usepackage{hyperref}       %
\usepackage{url}            %
\usepackage{booktabs}
\usepackage{tabularx}%
\usepackage{amsfonts}       %
\usepackage{nicefrac}       %
\usepackage{microtype}      %
\usepackage[dvipsnames]{xcolor} %
\usepackage[numbers,sort&compress]{natbib}
\usepackage{array,booktabs}
\usepackage{amsmath}
\usepackage{bbding}
\usepackage{txfonts}
\usepackage{pifont}
\usepackage{graphicx}
\usepackage{textcomp}
\usepackage{float}
\usepackage{rotating}
\usepackage{subcaption}
\usepackage{enumitem} %
\usepackage{wasysym} %

\usepackage{graphicx}
\newcommand {\otoprule}{\midrule [\heavyrulewidth]}
\newcolumntype {+}{ >{\global\let\currentrowstyle\relax}}
\newcolumntype {^}{ >{\currentrowstyle }}
 \newcommand {\rowstyle}[1]{\gdef\currentrowstyle{#1} %
 #1\ignorespaces
 }
\newcommand{\tabhead}{\rowstyle{\bfseries}}

\usepackage{xspace}
\newcommand{\modelname}{\textsc{XNNTab}\xspace}

\newcommand{\dsadult}{\textsc{AD}\xspace}
\newcommand{\dschurn}{\textsc{CH}\xspace}
\newcommand{\dscredit}{\textsc{CR}\xspace}
\newcommand{\dsspambase}{\textsc{SB}\xspace}
\newcommand{\dscovertype}{\textsc{CO}\xspace}
\newcommand{\dsgesture}{\textsc{GE}\xspace}
\newcommand{\dscar}{\textsc{CA}\xspace}
\usepackage{xcolor} %

\begin{document}
\title{\modelname -- Interpretable Neural Networks for Tabular Data using Sparse Autoencoders}
\author{Khawla Elhadri\orcidID{0009-0008-2696-8331} \and
Jörg Schlötterer\orcidID{0000-0002-3678-0390} \and
Christin Seifert\orcidID{0000-0002-6776-3868}}
\authorrunning{                                                                                                                       K. Elhadri et al.}
\institute{
Marburg University, Germany \\
\email{khawla.elhadri@uni-marburg.de, joerg.schloetterer@uni-marburg.de, christin.seifert@uni-marburg.de}
}
\maketitle              %
\begin{abstract}
In data-driven applications relying on tabular data, where interpretability is key, machine learning models such as decision trees and linear regression are applied. Although neural networks can provide higher predictive performance, they are not used because of their blackbox nature. In this work, we present \modelname, a neural architecture that combines the expressiveness of neural networks and interpretability. \modelname first learns highly non-linear feature representations, which are decomposed into monosemantic features using a sparse autoencoder (SAE). These features are then assigned human-interpretable concepts, making the overall model prediction intrinsically interpretable. \modelname outperforms interpretable predictive models, and achieves comparable performance to its non-interpretable counterparts. The code is available at \url{https://github.com/ke-pm/XNNTAB}.

\end{abstract}
\section{Introduction}
\label{sec:intro}
Tabular data is the most common type of data in a wide range of industries, including advertising, finance and healthcare.  While deep  neural networks (DNNs) achieved major advances in computer vision, their performance on tabular data remained below that of Gradient Boosted Decision Trees (GBDTs)~\citep{Chen_2016_XGBoostScalableTree, Khan_2022_TransformersVisionSurvey}. To improve the performance of DNNs on tabular data, several deep learning methods have been developed with specialized architectures that take into account the unique traits of tabular data (mixture of categorical, ordinal and continuous features) in order to be on par~\citep{Huang_2020_TabTransformerTabularData} or even outperform GBDTs~\citep{Chen_2024_CanDeepLearning}.
While these methods address the limitations of DNNs in terms of performance, DNNs still remain blackboxes, whose decisions are hard to communicate to end users~\citep{Schwalbe2023_ComprehensiveTaxonomyExplainable}.

In this paper, we address the interpretability limitations of existing DNNs and introduce \modelname\footnote{The research in this paper was presented in a early stage at UniReps: 3rd Edition of the Workshop on Unifying Representations in Neural Models at NeuRIPS 2025~\cite{elhadri2026interpretabledeepneuralnetworks}}, a deep neural network incorporating a sparse autoencoder (SAE). The SAE learns a dictionary of monosemantic, sparse features, and \modelname uses only these monosemantic features to generate outcome predictions, ensuring intrinsic interpretability. We take inspiration from the success of SAEs in learning interpretable features from LLM's internal activations~\citep{Huben_2023_SparseAutoencodersFind, yun_2021_transformervisualization}, and assume that monosemanticity holds for tabular data. While LLMs are used to explain the learned features in NLP,
we use an automated rule based method to assign human-understandable semantics to the learned dictionary features. Our method is illustrated in Figure~\ref{fig:method-overview} and described in Section~\ref{sec:method}.
We show that \modelname outperforms interpretable  models and matches the performance of blackbox models. \modelname is fully interpretable: the final prediction is a simple linear combination of dictionary features with clear human-understandable semantics.

\section{Related Work}
\label{sec:relatedwork}

Gradient boosted decision trees (GBDTs)~\citep{Chen_2016_XGBoostScalableTree,Khan_2022_TransformersVisionSurvey} have shown state-of-the art performance on tabular data~\citep{Grinsztajn_2022_Whytreebasedmodels}, outperforming standard neural architectures. 
To close this performance gap, neural architectures specifically tailored to tabular data have been proposed, such as attention-based architectures~\citep{Gorishniy_2021_RevisitingDeepLearning, Somepalli_2022_SAINTImprovedNeural, Yan_2023_T2GFORMERorganizingtabular}, and retrieval-augmented models~\citep{Somepalli_2022_SAINTImprovedNeural, Gorishniy_2023_TabRTabularDeepa}. Further approaches include regularizations~\citep{Jeffares_2023_TANGOSRegularizingTabular}, deep ensembles~\citep{Gorishniy_2024_TabMAdvancingtabular}, attentive feature selection~\citep{Arik_2021_TabNetAttentiveInterpretable}, feature embeddings~\citep{, Huang_2020_TabTransformerTabularData}, and reconstruction of binned feature indices~\citep{Lee_2024_Binningpretexttask}.

Recent methods focus not only on performance, but also aim to include interpretability.
ProtoGate~\citep{Jiang_2024_ProtoGateprototypebasedneural} is an ante-hoc interpretable architecture for tabular data, which learns prototypes of classes and predicts new instances based on their similarity to class prototypes. InterpreTabNet~\citep{Si_2024_InterpreTabNetdistillingpredictive}, a variant of TabNet~\citep{Arik_2021_TabNetAttentiveInterpretable} adds post-hoc explanations by learning sparse feature attribution masks and using LLMs to interpret the learned features from the masks.
Similar to ProtoGate and different from InterpetTabNet,  \modelname is intrinsically interpretable, i.e., it uses interpretable features directly for prediction. While ProtoGate relies on class prototypes, \modelname learns relevant feature combinations that describe parts of instances, similar to part-prototype models~\citep{elhadri_2025_lookslikewhatchallenges}.

\section{\modelname}
\label{sec:method}

\begin{figure*}
    \centering
    \includegraphics[width=\linewidth]{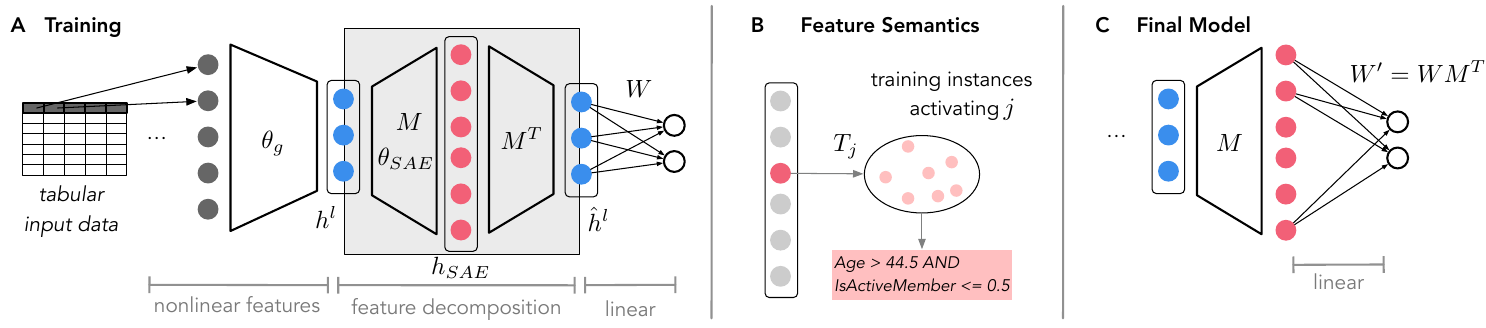}
    \caption{Architecture overview. \textbf{A.} An MLP learns nonlinear features, which are decomposed into monosemantic dictionary features using an SAE. \textbf{B.} Dictionary features are assigned a human-interpretable meaning by learning rules for the subset of training instances that highly activate a specific feature. \textbf{C.} Predictions are linear combinations of monosemantic features by combining the linear model components (the decoder, $M^T$, and the linear layer of the MLP, $W$).}
    \label{fig:method-overview}
\end{figure*}

\modelname consists of a standard neural blackbox (e.g., an MLP) to learn nonlinear latent features, and a sparse autoencoder (SAE)~\citep{Huben_2023_SparseAutoencodersFind} to decompose this latent polysemantic feature space into monosemantic features (cf. Figure~\ref{fig:method-overview}, A). The monosemantic features are in a latent space with unknown (but unique) semantics. To assign meaning to each feature, we learn rules that describe the subset of the training samples that highly activate each feature (cf. Figure~\ref{fig:method-overview}, B). Finally, we combine the last linear layers into a single linear layer by multiplying their weight matrices (cf. Figure~\ref{fig:method-overview}, C), making the final prediction a direct linear combination of the monosemantic, rule-defined features.. 

\subsection{Architecture}
The base blackbox model is an MLP with multiple hidden layers and a softmax output. We denote the penultimate layer as $l$, and the representations of this layer as $h_l$.  
The MLP learns a function $\hat{y} = f(x)$. $f(x)$ can be decomposed into a function $g$ learning the hidden representations  $h_l$ and a linear model $h$ predicting the target based on $h_l$, i.e., $\hat{y} = h(h_l)=h(g(x))$. $g(\cdot)$ is characterized by the parameters $\theta_g$, and $h(\cdot)$ is characterized by the weight matrix $W$.

We adopt the SAE architecture described in~\cite{Huben_2023_SparseAutoencodersFind}, consisting of an input layer, a hidden layer, and an output layer. The weights of the encoder and decoder are shared, and given by the matrices $M$ and $M^T$, respectively. The encoder has an additional bias vector $b$, the decoder has not~\citep{Huben_2023_SparseAutoencodersFind}. We set the size of the hidden layer as $d_{hid} = R\cdot d_{in}$ with $d_{in}$ being the dimension of the MLP's layer $l$. $R$ is a hyperparameter that controls the size of the feature dictionary.
The input to the SAE are the hidden representations of the blackbox MLP $h_l$.  The output of the SAE is given by 
$\hat{h}_l = M^T \Bigl(ReLU(M h_l + b)\Bigr)$.
By design, $\hat{h}_l$ is a sparse linear combination of (latent) monosemantic dictionary features~\citep{Huben_2023_SparseAutoencodersFind}.

\subsection{Training}
\modelname is trained in four steps: i) training of the MLP's representation, ii) training of the SAE, iii) finetuning of the decision layer, and iv) combining linear components. 

\textbf{Step 1: MLP training.} The MLP's parameters $\theta_\text{MLP}$= ($\theta_g$, $W$) are trained using cross-entropy loss and L1 regularization. The SAE is not present in this step.

\textbf{Step 2: SAE Training.} To learn a dictionary of monosemantic features we follow the training procedure of \cite{Huben_2023_SparseAutoencodersFind}. The SAE is trained to reconstruct the hidden activations $h_l$ of training samples $x$. The SAE uses a reconstruction loss on its output and a sparsity loss on its hidden activations $h_{SAE}$ to learn sparse dictionary features.

\textbf{Step 3: Finetuning the decision layer.} 
We freeze the parameters $\theta_g$ of the MLP and $\theta_{SAE}$ of the SAE and finetune the weights $W$ in the decision layer with the same loss as in step 1 to make predictions from the reconstructions $\hat{h}_l$ learned by the SAE.

\textbf{Step 4: Aggregating linear components.} Both, the decoder part of the SAE and the final decision layer of the MLP are linear layers, and can be combined into one layer. 
Therefore, we directly connect the hidden layer of the SAE to the output and set the weights of this layer to $W' = WM^T$.\footnote{$\hat{h}_l = M^T h_{SAE}$, and $\hat{y} = W \hat{h}_l = W M^T h_{SAE}$ }

\subsection{Learning Representation Semantics}
The hidden representations of the SAE represent monosemantic dictionary features in a latent space, but their semantics are unknown. To assign human-understandable semantics to those features, we follow a similar idea as in~\citep{Huben_2023_SparseAutoencodersFind} for tabular data instead of text (cf. Figure~\ref{fig:method-overview}, B). 
For each dimension in the latent representation of the SAE $j \in \{1, ...d_{hid} \}$, we identify the subset of training samples $T_j$ which highly activate this feature, i.e., whose activations are above a threshold $p$. We then use a rule-based classifier to learn simple decision rules describing this subset.

\section{Experimental Setup}
\label{sec:experiments}
We train \modelname on 7 publicly available datasets, and compare the performance to interpretable models, classical blackbox models and state-of-the art deep learning models. 

\subsection{Datasets and Evaluation Metrics}
\label{ssec:exp:data-eval}
We evaluated \modelname on benchmark datasets for structured data from previous literature: Adult~\cite{adult_dataset}, Churn Modeling\footnote{https://www.kaggle.com/shrutimechlearn/churn-modelling}, Spambase\footnote{https://openml.org/search?type=data\&status=active\&id=44}, Credit\_g\footnote{https://openml.org/search?type=data\&status=active\&id=31}, Gesture Phase Prediction~\cite{gesture_dataset}, Covertype~\cite{covertype_dataset} and Car\footnote{https://openml.org/search?type=data\&status=active\&id=40975}. An overview of the datasets is shown in Table~\ref{tab:dataset}. We used 5-fold-cross-validation with 65\%/15\%/20\% (training/validation/test) splits. We report macro F1 on the test set, averaged across all folds.

\begin{table}[tb]
\centering
\caption{Dataset overview. Showing the number of instances, the number of numerical features (\# Num.), the number of categorical features (\# Cat.) and the machine learning task (Task).}
\label{tab:dataset}
\begin{tabular}{@{\extracolsep{10pt}}+c^c^c^c^c^c}
\toprule\tabhead
Abbr & Name & Instances & \# Num. & \# Cat. & Task \\ \midrule
\dsadult & Adult  & 48,842 &   6 & 9  & binary classification \\
\dschurn & Churn Modeling & 10,000 & 10 & 1& binary classification \\
\dscredit & Credit\_g & 1,000 & 7 & 14 & binary classification \\
\dsspambase & Spambase & 4,601 & 57 & 1& binary classification \\
\dsgesture & Gesture & 9,873 & 32 & 0 & multi-class classification \\
\dscovertype & Covertype & 581,012 & 54 & 0 & multi-class classification \\
\dscar & Car & 1,728 & 0 & 7 & multi-class classification \\
\bottomrule
\end{tabular}
\end{table}

\subsection{Models and Parameters}
\label{ssec:exp:models}
We compare \modelname to interpretable baselines, classical machine learning models and neural models, which are specifically designed for learning on tabular data.

As baselines, we use two interpretable models, Logistic Regression (LR) and Decision Trees (DTs), and two blackbox models, XGBoost and Random Forests (RF). We further compare \modelname to recent deep learning models, namely TabNet~\cite{Arik_2021_TabNetAttentiveInterpretable}, NODE~\cite{popov_2019_NODE}, T2G-Former~\cite{Yan_2023_T2GFORMERorganizingtabular}, FT-Transformer~\cite{Gorishniy_2021_RevisitingDeepLearning}, TabM~\cite{Gorishniy_2024_TabMAdvancingtabular}, TabR~\cite{Gorishniy_2023_TabRTabularDeepa} and InterpreTabNet~\cite{Si_2024_InterpreTabNetdistillingpredictive}. We report the experimental results of each model from their respective papers, TabNet and Node results are reported from T2G-Former. 
Note, that TabNet, NODE, T2G-Former, FT-Transformer, TabM, TabR and InterpreTabNet  use a different training process: 
each dataset is split once into  65\%/15\%/20\% (training/validation/test). After tuning hyperparameters on the validation set, 15 experiments are run with different random seeds, reporting average performance on the test set.

Our neural base model for \modelname is an MLP, with a different architecture for each dataset. 
We perform a neural architecture search on each cross-validation split and select the architecture with the smallest last hidden layer overall. We use the validation sets for tuning other hyperparameters using the Optuna library~\cite{optuna_2019}. See Appendix~\ref{appendix:impl_details:paramter_settings} for details on all hyperparameters.

\subsection{Learning Representation Semantics}
\label{ssec:exp:rule-extractor}

To assign human-understandable semantics to the dictionary features, we employ Skope-rules\footnote{https://github.com/scikit-learn-contrib/skope-rules. Accessed July 2025}. 
The Skope-rules classifier learns rule-based descriptions of a target subset by framing the task as binary classification: instances in the subset are labeled 1, all others 0. It outputs candidate rules along with their recall, each capturing different portions of the subset. If no suitable rule exists, it may return an empty rule set.

We identify the instances that most strongly activate each dictionary feature and use Skope-rules to describe these features. 
We use the default Skope-rules hyperparameters and set the precision to $1.0$.
Preliminary experiments showed that no single activation threshold yields rules for all features: some neurons never activate, some rarely activate and some produce no extractable rule at certain thresholds. However, all activated features are used for prediction. To obtain an optimal rule set for the dictionary features $f_j$, we therefore perform the following steps: 

\begin{enumerate}[topsep=0pt, itemsep=0pt]
    \item Pass the dataset through the model and record each neuron's activation, yielding one activation value per instance.
    \item For $p \in \{90\%, 80\%, 70\%, 60\%, 50\%\}$, select the top-$p$ activations of neuron $j$ and the corresponding instances, forming the subset $T^p_j$.
    \item For each $T^p_j$, label its instances as $1$ and all others as $0$, run Skope-rules, and select the rule with the highest recall.
    \item For each $p$, count how many dictionary features obtained a rule and choose the $p$ that maximizes this count; break ties using the highest average recall.
\end{enumerate}

\section{Results}
\label{sec:exp:results}
In this section, we report predictive performance, show qualitative and quantitative results on interpretability and analyze the impact of the activation threshold. 

\subsection{Performance}
\label{sec:exp:results:performance}

\begin{table}[tbp]
\centering
\caption{Macro F1 of blackbox (\ding{53}) and fully interpretable (\checkmark) models on two benchmark datasets. Best values are marked bold, best values for interpretable models bold italic. The standard deviation for all experiments is small (max: 0.032, min: 0.001).}
\label{tab:results:performance-core}
\begin{tabular}{c@{\hspace{6pt}}l@{\hspace{6pt}}c@{\hspace{6pt}}c@{\hspace{6pt}}c@{\hspace{6pt}}c@{\hspace{6pt}}c@{\hspace{6pt}}c@{\hspace{6pt}}c}
\toprule
 &   & \multicolumn{1}{c}{\dsadult $\uparrow$}  & \multicolumn{1}{c}{\dschurn $\uparrow$} & \multicolumn{1}{c}{\dscredit $\uparrow$} & \multicolumn{1}{c}{\dsspambase $\uparrow$} & \multicolumn{1}{c}{\dscovertype $\uparrow$} & \multicolumn{1}{c}{\dsgesture $\uparrow$} & 
 \multicolumn{1}{c}{\dscar $\uparrow$}\\ 
 \otoprule
\ding{53} & RF 
    & 0.799  & 0.747  & 0.657  & 0.947 & 0.838  & 0.596  & 0.908 \\
\ding{53} & XGBoost 
    & \textbf{0.815} &  0.750  & \textbf{0.716} & \textbf{0.957}  &\textbf{ 0.942}  & \textbf{0.638}  &\textbf{0.974}\\
\ding{53} & MLP 
    & {0.796} &  0.751  & 0.693 & 0.945 & 0.870 & 0.632  & 0.955\\\midrule
\checkmark & LR 
    & 0.782  &  0.601 & \textit{\textbf{0.701}}  &  0.921 & 0.476  & 0.224 & 0.358 \\
\checkmark & DTs
    & 0.787  &  0.735  &  0.645 &  0.915  & 0.844  & 0.448  &  0.932 \\
\checkmark  & \modelname
    & \textit{\textbf{0.795}} &  \textit{\textbf{0.752}}& 0.700  & \textit{\textbf{0.948}} & \textit{\textbf{0.878}} &\textit{\textbf{0.634}} & \textit{\textbf{0.955}}\\

\bottomrule
\end{tabular}
\end{table}

Table~\ref{tab:results:performance-core} shows that,  on nearly all datasets, \modelname outperforms interpretable models, with the exception of \dscredit, for which \modelname is on par with Logistic Regression. 
On \dsadult, \modelname is comparable to MLP and slightly below the performance of XGBoost and Random Forest. On \dschurn and \dsgesture, \modelname is on par with all baseline blackbox models including XGBoost.
On \dscredit, \dsspambase and \dscovertype, \modelname outperforms MLP and is slightly below XGBoost. These results show that our model introduces interpretability to DNNs with very little compromise on performance.

\begin{table}[th]
\centering
\caption{Accuracy of related work models in comparison to \modelname. Blackbox models are marked as (\ding{53}), partially interpretable models as $\ocircle$, and fully interpretable models as (\checkmark) models. Standard deviation for all experiments is at most 0.144 (on \dsgesture by T2G-Former), at least 0.000. Results for other models are taken from the respective papers, n.a. indicates that values are not available. Note: 
Related work models follow a different training pipeline than ours (see Sec.~\ref{ssec:exp:models}).} 
\label{tab:results:performance-relwork}
\begin{tabular}{c@{\hspace{6pt}}l@{\hspace{6pt}}c@{\hspace{6pt}}c@{\hspace{6pt}}c@{\hspace{6pt}}c@{\hspace{6pt}}c@{\hspace{6pt}}c@{\hspace{6pt}}c}

\toprule
 &  & \dsadult$\uparrow$ & \dschurn$\uparrow$ &  \dscovertype$\uparrow$ &
\dsgesture$\uparrow$ \\
\midrule
\ding{53} & TabM
    & 0.857  & 0.860  &  0.973  & n.a \\
\ding{53} & TabR
    & 0.870  & 0.862  & 0.976  & n.a \\
\ding{53} & FT-Transformer 
    & 0.859 & 0.860 &  0.970 & 0.613 \\

\ding{53} & NODE
    & 0.857 & 0.858 &  n.a & 0.539\\
\ding{53} & T2G-FORMER
    & 0.859 & 0.862 & n.a & 0.655\\
\midrule
$\ocircle$ & TabNet
    & 0.848 & 0.850 &  0.957 & 0.600\\
$\ocircle$& InterpreTabNet
 & 0.874 & n.a &  n.a & n.a\\
 \midrule
\checkmark & \modelname
    & 0.850 & 0.861 &  0.968 & 0.665\\
\bottomrule
\end{tabular}

\end{table}

Table~\ref{tab:results:performance-relwork} shows accuracy of \modelname in comparison to state-of-the art neural models. Note that \modelname is trained using 5-fold-cross-validation, while the other methods only use one split (with the data setup not reported in the literature) with 15 random seeds (cf. Sec.~\ref{ssec:exp:models}).
\modelname scores higher than TabNet on all datasets where results are available in the literature. 
On \dsadult, \modelname performs comparably to TabM, NODE, FT-Transformer and T2G-FORMER and underperforms TabR and the partially interpretable InterpreTabNet. On \dschurn, \modelname is comparable to all blackbox models. On \dscovertype, \modelname has lower performance than TabR and is comparable to the other blackbox models. On \dsgesture, \modelname outperforms all models under comparison. These results indicate that \modelname has little compromise on performance compared to the state-of-the-art while being interpretable, and can be an alternative to blackbox models.

\subsection{Interpretability}
\label{sec:exp:results:interpretability}

\begin{figure}
    \centering
    \includegraphics[width=0.9\linewidth]{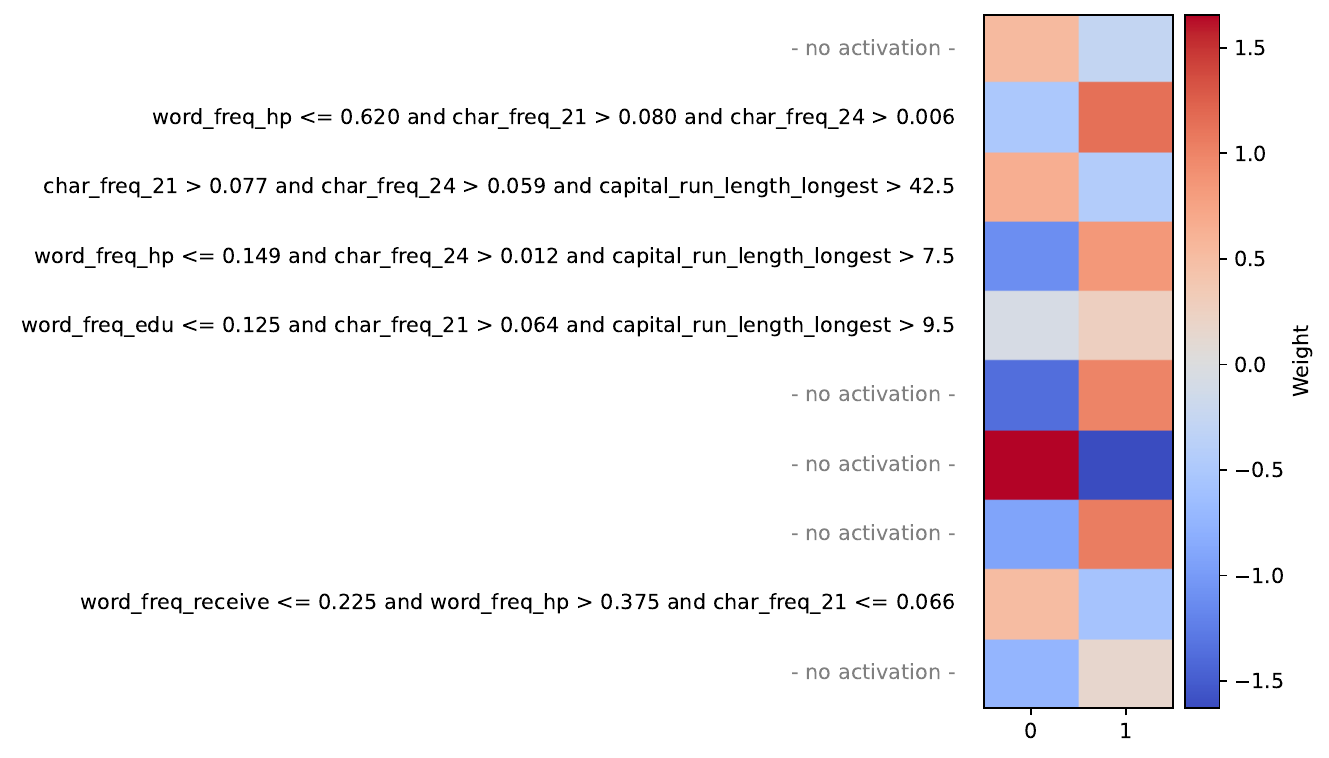}
    \caption{Heatmap of decision weight matrix $W'$ for \dsspambase on interpretable features.}
    \label{fig:weight_matrix}
\end{figure}

 A selection of rules extracted for the dictionary features on \dsadult are shown in Table~\ref{tab:dict_features_adult_selection} and for \dsspambase in Table~\ref{tab:dict_features_spambase} (full list).
The learned rules provide a global explanation of the model's behavior (Figure~\ref{fig:weight_matrix}), since the predictions are simply a linear combination of the learned features (rules). Each weight in this linear combination directly indicates whether the presence of the corresponding features increases or lowers the probability of a class, and to which extent.

Table~\ref{tab:results:global-expl-all-dataset} shows the average number of active features for each dataset. For \dscredit, 19 out of the 57 features that we learn are active for the chosen $p$ (70\%).
That means, one local explanation requires us to inspect 19 rules on average to understand the model's prediction. However, that number decreases to 3 for \dsspambase with a threshold of $p = 80 \%$.
While the average number of active features varies for each dataset, the complexity of the rules is on average 2.4 for all datasets. That means, each rule is simple to understand, as it has approximately two conditions on average.

\begin{table}[tbhp]
    \centering
    \scriptsize
    \caption{A selection of dictionary features for the \dsadult dataset. $|T_j|$ - size of the training subset that strongly activate feature $j$. Coverage of the rule reported as number of samples and percentage of samples. Table sorted by $|T_j|$.}
    \label{tab:dict_features_adult_selection}
    \begin{tabularx}{\textwidth}{l l X l}
    \toprule
     j & |T\_j| & Description & Coverage \\
    \midrule
    40 & 15294 & \texttt{marital\_status\_Married is False and educational\_num < 13 and capital\_gain <= 8028.0} & 11625/0.76 \\
     9 & 11968 & \texttt{marital\_status\_Married is False and age <= 37.5 and educational\_num < 12} & 7271/0.61 \\
     8 & 11634 & \texttt{marital\_status\_Married is False and age <= 34.5 and educational\_num < 12} & 6517/0.56 \\
    54 & 10889 & \texttt{marital\_status\_Married is False and age <= 31.5 and educational\_num < 13} & 5953/0.55 \\
    44 &  7457 & \texttt{marital\_status\_Married is False and relationship\_Not\_in\_family is False and age <= 25.5} & 3330/0.45 \\
    45 &  3975 & \texttt{marital\_status\_Married is False and age <= 22.5 and hours\_per\_week <= 32.5} & 1516/0.38 \\
     5 &  1758 & \texttt{age <= 20.5 and hours\_per\_week <= 24.5} & 754/0.43 \\
    34 &  1658 & \texttt{occupation\_Other\_service is False and capital\_gain > 14682.0} & 547/0.33 \\
    26 &  1358 & \texttt{marital\_status\_Widowedand is False and capital\_gain > 14682.0} & 546/0.40 \\
    50 &   982 & \texttt{age <= 72.0 and capital\_gain > 15022.0} & 532/0.55 \\
    63 &   978 & \texttt{gender is Female and age <= 18.5 and hours\_per\_week <= 24.5} & 236/0.24 \\
     7 &   290 & \texttt{occupation\_Farming\_fishing is False and capital\_gain > 19266.0 and hours\_per\_week > 27.5} & 209/0.73 \\
    18 &   226 & \texttt{relationship\_Other\_relative is False and capital\_gain > 26532.0} & 185/0.82 \\
    22 &   225 & \texttt{occupation\_Farming\_fishing is False and capital\_gain > 26532.0 and hours\_per\_week > 18.0} & 183/0.82 \\
     3 &   212 & \texttt{capital\_gain > 34569.0} & 153/0.73 \\

    \bottomrule
\end{tabularx}
\end{table}

\begin{table}[tbhp]
    \centering
    \scriptsize
    \caption{Dictionary features for the \dsspambase dataset. $|T_j|$ - size of the training subset that strongly activate feature $j$. Coverage of the rule reported as number of samples and percentage of samples. Table sorted by $|T_j|$.}
    \label{tab:dict_features_spambase}
\begin{tabularx}{\textwidth}{+l^l^X^l}
\toprule
j & |T\_j| & Description & Coverage \\
\midrule
5 &  2340 
    & \texttt{word\_freq\_edu <= 0.125 and char\_freq\_21 > 0.0645 and capital\_run\_length\_longest > 9.5} 
    & 984/0.42 \\
4 & 2240 
    & \texttt{word\_freq\_hp <= 0.14998 and char\_freq\_24 > 0.0119 and capital\_run\_length\_longest > 7.5} 
    & 693/0.31 \\
9 &  1415 
    & \texttt{word\_freq\_receive <= 0.2249 and word\_freq\_hp > 0.375 and char\_freq\_21 <= 0.0659}
    & 465/0.33 \\
2 &  1031 
    &\texttt{word\_freq\_hp <= 0.6200 and char\_freq\_21 > 0.0804 and char\_freq\_24 > 0.0064}
    & 586/0.56 \\
3 &   833 
    & \texttt{char\_freq\_21 > 0.0775 and char\_freq\_24 > 0.0595 and capital\_run\_length\_longest > 42.5} & 332/0.40 \\
\bottomrule
\end{tabularx}
\end{table}

\begin{table}[tbhp]
\centering
    \caption{Complexity of features on all datasets showing rule length (number of single terms in a rule), and active features per decision.}
        \label{tab:results:global-expl-all-dataset}
    \begin{tabular}{@{\extracolsep{10pt}}^l+l+c+c+c+c+c+c+c}
    \toprule\tabhead
    & & \dsadult & \dschurn & \dscredit & \dsspambase & \dscovertype& \dsgesture& \dscar\\\midrule
    Rule Length  
    & mininum & 1   & 1 &2 &3 & 1 & 2 & 1 \\
    &average  & 2.18 & 1.8 & 2.9&3 & 2.2 &2.8 & 2.2\\
    &maximum  & 3   & 2 & 3 & 3 & 3 & 3 &3\\\midrule
    Active Features 
    &mininum & 23   & 19 & 13 & 1 & 89 & 25 & 21 \\
    &average & 28.7 & 33 & 19 & 3.3 & 115 & 32& 27 \\
    &maximum & 34  & 40  & 23 & 5 & 137 & 40 & 35\\\bottomrule
    \end{tabular}  
\end{table}

\subsection{Ablation on Activation Threshold}
\label{sec:exp:results:ablation}
The choice of $p$ influences model interpretability in two ways. First,  different $p$ result in a different amount of features for which rules can be found, i.e., it defines the extend to which the model's reasoning is explained. Second, different $p$ result in different $|T_j|$, and therefore different recall values for the rules. The recall describes how many instances of $|T_j|$ are described by the rule and is therefore a measure of reliability.

\begin{figure}[tbhp]
    \centering
    \begin{subfigure}[b]{0.65\textwidth}
        \centering
        \includegraphics[width=\textwidth]{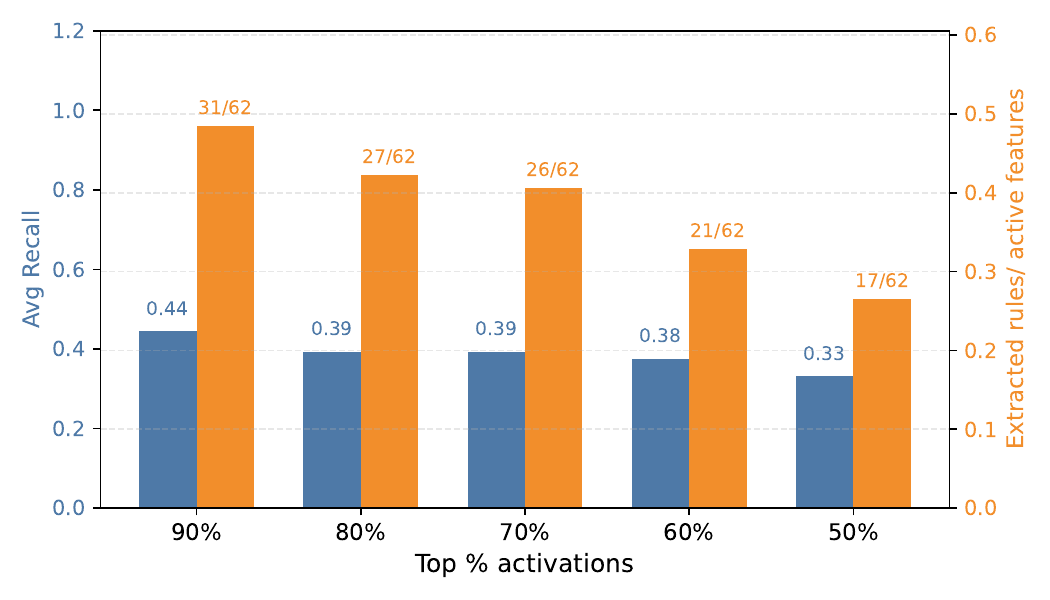}
        \subcaption{\dsgesture dataset}
    \end{subfigure}
        \hfill
     \begin{subfigure}[b]{0.65\textwidth}
    \includegraphics[width=\textwidth]{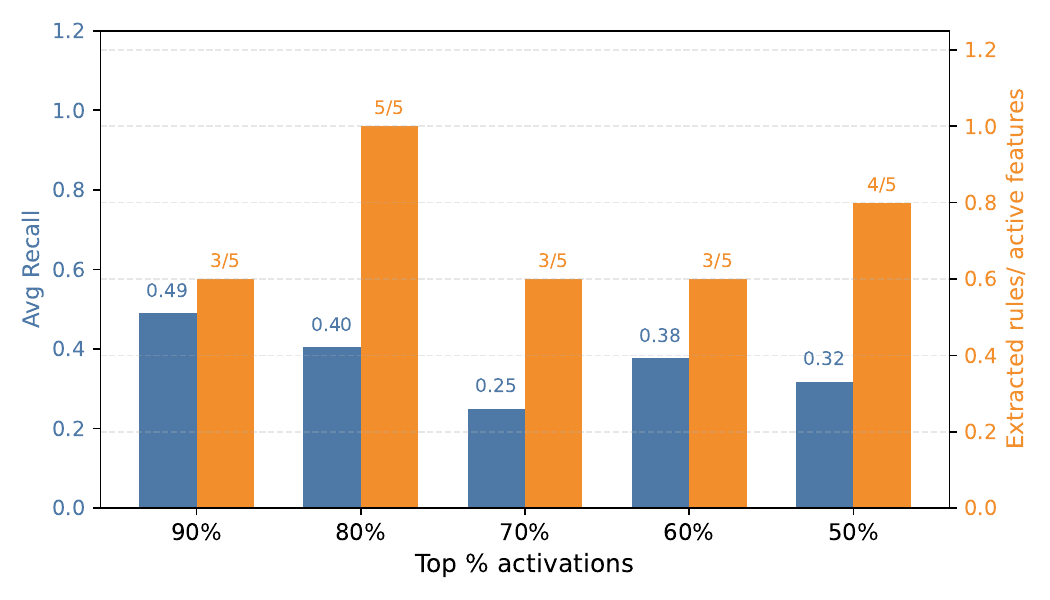}
    \subcaption{\dsspambase dataset}
    \end{subfigure}
    \caption{Feature coverage on \dsgesture dataset (top) and \dsspambase dataset (bottom) for different $p$.
    \textbf{Orange bars} show the proportion of active features for which a rule was successfully extracted, \textbf{blue bars} show the average recall of the extracted rules.}
   \label{fig:thresholds}
\end{figure}

Figure~\ref{fig:thresholds} shows for how many of the active features rules can be extracted, and the fraction of instances covered by rules (recall) for $p \in \{50\%,60\%,70\%,80\%,90\%\}$.
On \dsgesture, lower $p$ lead to fewer features being covered by Skope-rules, their average recall decreases accordingly. However, this behavior does not generalize over all datasets; for \dsspambase, both the number of extracted features and their respective average recall fluctuate as $p$ decreases. Ultimately, choosing an optimal  $p$ for interpretability seems to depend on the dataset and requires to trade-off the number of covered features (i.e., the extent to which the model can be explained) and the average rule recall (i.e., reliability of a rule).

\section{Conclusion}
In this paper, we introduce \modelname, an interpretable model for tabular data. Our results show, that \modelname outperforms interpretable models and performs comparably to blackbox architectures across several datasets, while being interpretable. By using a rule-based classifier, we show how the model's output predictions can be explained by simple decision rules. 

Although our model is interpretable, it has limitations: learning decision rules to explain learned features is post-hoc and not always complete, as it is not always possible to extract rules for all features. Furthermore, a high number of learned features results in a high average number of active features per prediction, which limits interpretability.

Directions for future work include investigating strategies to reduce redundant rules such as pruning or combining neurons, and exploring automated approaches for selecting activation thresholds that balance rule recall with the fraction of extracted rules. The SAE approach could also be extended to other blackbox neural models for tabular data to further enhance performance.

\begin{credits}
\subsubsection*{\ackname}
This work was funded by the Deutsche Forschungsgemeinschaft (DFG, German Research Foundation) under project ID 536124560.
\end{credits}
\appendix
\section{Implementation Details}
\subsection{Datasets}

Standard preprocessing was applied to all datasets: removal of non-informative features (e.g. ``RowNumber'', ``CustomerId'), min-max normalization of numerical features, one hot encoding of nominal variables, and ordinal encoding of ordinal variables.

\subsection{Hyperparameters}
\label{appendix:impl_details:paramter_settings}
For all models, we use Optuna~\footnote{\url{https://optuna.org}} for hyperparameter tuning. We tune on the validation set of each split (5 splits in total). For each experiment, the number of iterations is set to 100. Random seed = 0.

\textbf{RF.} We tuned $n\_estimators$ in [100, 200, 300], $max\_depth$ in [5, 10, 15, 20], $min\_samples\_split$ in [2, 10], and  $min\_samples\_leaf$ in [1, 10].

\textbf{XGBoost.} We set the booster to \textit{gbtree}, $early\_stopping\_rounds$ to 12, and tuned $learning\_rate$ in [0.001, 0.4],  $max\_depth$ in [3, 10], $subsample$ in [0.5, 1], $\lambda$ in  [0.1, 10]

\textbf{DT.} We tuned  $max\_depth$ in [5, 10, 15, 20],  $min\_samples\_split$ in [2, 10], and  $min\_samples\_leaf$ in [1, 10].

\textbf{LR.} We tuned $max\_iter$ in [100, 200].

\textbf{MLP.} We tuned the learning rate ($lr$) in  [$1e-2$, $5e-3$], $Dropout$ in [0.0, 0.5] and  $\lambda$ in [$1e-7$, $1e-2$] for L1 regularization.
 
\textbf{\modelname.} Additional to the MLP parameters, SAE parameters are the $L_1$ coefficient $\alpha = 1e-3$ for \dsadult and \dschurn.

\textbf{Skope-rules.} We set $precision\_min = 1$, $recall\_min = 0.2$, and used the default settings for the rest of the parameters.

\paragraph{Model Architectures:}
\begin{small}
\begin{itemize}
    \item \dschurn and \dsadult: 3-layer base MLP (100, 64, 32);  SAE with $2 \times 24 = 64$ (R=2) neurons.
    \item \dscredit: 3-layer base MLP (174, 180, 19); SAE $3 \times 19 = 57$ (R=3) neurons.
    \item \dsspambase: 3-layer base MLP (96, 179, 5); SAE $2 \times 5 = 10$ (R=2) neurons.
    \item \dsgesture: 3-layer base MLP(128, 64, 32);  SAE $2 \times 32 = 64$ (R=2) neurons.
    \item \dscovertype: 2-layer base MLP (169, 175), SAE $1 \times 175 = 175$ (R=1) neurons.
    \item \dscar: 2-layer base MLP (106, 44): the SAE $2 \times 44 = 88$ (R=2) neurons.
\end{itemize}
\end{small}

\bibliographystyle{splncs04}
\bibliography{xai.bib}
\end{document}